\documentclass{article}
\usepackage{arxiv}

\usepackage{times}
\usepackage{soul}
\usepackage{url}
\usepackage[hidelinks]{hyperref}
\usepackage[utf8]{inputenc}
\usepackage[small]{caption}
\usepackage{graphicx}
\usepackage{amsmath}
\usepackage{amssymb}
\usepackage{amsfonts}
\usepackage{amsthm}
\usepackage{booktabs}
\usepackage{array}
\usepackage[table]{xcolor}
\usepackage{subcaption}
\usepackage[ruled,vlined]{algorithm2e}
\usepackage[switch]{lineno}
\usepackage{natbib}

\newtheorem{definition}{Definition}

\title{DeFed-GMM-DaDiL: A Decentralized Federated Framework for Domain Adaptation}

\author{
Rebecca Clain\\
Universit\'e Paris-Saclay, CEA, List, F-91120 Palaiseau, France\\
\texttt{clain.rebecca@hotmail.fr}
\And
Eduardo Fernandes Montesuma\\
Sigma Nova\\
\texttt{edumontesuma@gmail.com}
\And
Fred Ngol\`e Mboula\\
Universit\'e Paris-Saclay, CEA, List, F-91120 Palaiseau, France\\
\texttt{fred-maurice.ngole-mboula@cea.fr}
}

\begin{document}

\maketitle

\begin{abstract}
Decentralized multi-source domain adaptation seeks to transfer knowledge from multiple heterogeneous and related source domains to an unlabeled target domain in a decentralized setting. We address this challenge through a fully decentralized federated approach, DeFed-GMM-DaDiL, an extension of the GMM-Dataset Dictionary Learning (DaDiL) framework. Each client models its dataset as a Gaussian Mixture Model (GMM), and the federation jointly approximates them via labeled Wasserstein barycenters of shared, learnable GMM atoms. This design enables adaptation without a central server while preserving clients' privacy. We empirically study the stability of the learned representations in scenarios where the target domain has missing classes. Empirical results demonstrate that DeFed-GMM-DaDiL maintains stable and consistent shared representations across clients, effectively reconstructs missing classes, and achieves competitive performance on multi-source domain adaptation benchmarks.
\end{abstract}

\section{Introduction}


Machine learning models are typically developed under the assumption that data are identically distributed, yet in practice, data often come from different domains with varying distributions. These distributional differences can lead to severe performance degradation \cite{quinonero2008dataset}. Domain adaptation provides a framework for addressing this challenge. In particular, Multi-Source Domain Adaptation (MSDA) leverages labeled data from multiple heterogeneous source domains to improve generalization to an unlabeled target domain. In distributed settings, where data cannot be centralized due to privacy or communication constraints, decentralized MSDA methods enable adaptation across multiple clients. However, most decentralized MSDA approaches still depend on a central server to aggregate model updates, introducing bottlenecks and single points of failure. Eliminating this dependency is critical for robust and scalable decentralized MSDA. 
An additional challenge in MSDA arises from the partial class coverage in the target domain. In applications such as defect detection or rare disease diagnosis, the target client may not observe all classes, resulting in an incomplete distribution that can degrade adaptation performance. Most decentralized MSDA methods do not address this issue, highlighting the need for approaches capable of reconstructing or transferring knowledge for unseen classes. To address these challenges, we propose DeFed-GMM-DaDiL, a fully decentralized federated approach based on the GMM-DaDiL framework \cite{GMMDADIL}. GMM-DaDiL represents each domain as a Gaussian Mixture Model (GMM) and approximates it through a Wasserstein barycenter of a shared, learnable dictionary of atoms, enabling adaptation to distributional shifts. We adapt this framework to a fully federated setting without a central server and demonstrate that it achieves strong performance while being robust to missing classes.
The remainder of this paper is organized as follows. Section \ref{sec:related_work} reviews related work on MSDA. Section \ref{sec:background} introduces the necessary background. Section \ref{sec:proposed_approach} presents the proposed DeFed-GMM-DaDiL framework.
Section \ref{sec:missing_classes} analyzes the stability of the proposed approach in the presence of missing target classes.  Finally, Section \ref{sec:conclusion} concludes the paper.

\section{Related work}
\label{sec:related_work}
\subsection{Optimal transport for MSDA}
In MSDA, Optimal Transport (OT), a field of mathematics concerned with the displacement of mass at minimal cost, provides a rigorous framework for comparing and relating distributions across domains. In particular, OT has been used either to adjust the contribution of each source domain through reweighting \cite{turrisi2022multisourcedomainadaptationweighted} or to aggregate the source domains into a Wasserstein barycenter and map it to the target \cite{Montesuma2021acoustic,Montesuma2021DA}. Following this line of work, \cite{Montesuma2023} represent domains as barycenters of free distributions with learnable supports, enabling interpolation of distributional shifts. While effective, barycenter-based methods \cite{Montesuma2021acoustic,Montesuma2021DA,Montesuma2023} scale poorly because empirical supports grow with dataset size. To address this issue, \cite{GMMDADIL} propose a parametric approach based on optimal transport between Gaussian Mixture Models (OT-GMM) \cite{delon}, allowing domains to be represented compactly and barycenters to be computed efficiently. Building on this OT–GMM framework, our contribution extends it to a fully decentralized and federated setting, where domains collaborate directly without any central server or coordination. This decentralization is crucial for scenarios where data cannot be centralized due to privacy, communication, or organizational constraints.

\subsection{Decentralized MSDA}
Although OT-based MSDA frameworks effectively capture distributional shifts, most approaches remain centralized and have not been applied in a fully federated setting. In distributed settings, decentralized MSDA approaches adapt models from multiple heterogeneous source domains to an unlabeled target without centralizing data. Existing literature on decentralized MSDA includes the work of \cite{peng2019federated}, which aligns representations across nodes with the target distribution through adversarial learning and feature disentanglement; the work of \cite{liu2023co}, which treats source nodes as black-box models that share only soft outputs (the probabilities assigned by source models to each class); and that of \cite{feng2021kd3a}, which leverages knowledge distillation across source models. Despite their emphasis on privacy, these approaches still depend on a central server for parameter aggregation, which introduces potential security vulnerabilities. Centralized aggregation creates a single point of failure and exposes clients to privacy inference risks, as sensitive information can be partially reconstructed from shared model updates. For instance, \cite{feng2021kd3a} relies on exchanging model parameters through a central coordinator, while \cite{liu2023co} mitigates data sharing by transmitting only soft labels, but both approaches ultimately rely on centralized aggregation.
In this work, we propose a fully serverless approach to decentralized MSDA, enhancing robustness while mitigating privacy risks associated with central aggregation.

\subsection{MSDA missing classes challenge}
Another key challenge in MSDA is partial class coverage in the target domain, where some classes are missing from the available target data despite being present in the underlying distribution. Ignoring these classes can lead to misalignment and negative transfer — i.e. harmful knowledge transfer that degrades adaptation performance. Previous work has shown that standard MSDA methods fail in such cases \cite{wangmissingclasses}, and several studies have emphasized the importance of addressing this issue. For instance, \cite{Jing2021TowardsNT} reduces the risk of negative transfer by selectively aligning target samples with relevant source neighborhoods. \cite{ishiimissingclasses} tackle partially missing target classes by using instance weighting to estimate the contribution of unseen classes, allowing the model to consider all classes during training. Eventually, \cite{wangmissingclasses} highlights how disentangling domain-invariant and class-specific features can improve robustness. These works illustrate that dealing with incomplete target label spaces is a crucial requirement for a realistic and robust MSDA. Positioned within this line of research, we evaluate our approach under missing-class scenarios and show that it remains robust despite varying class coverage across source and target domains.
\section{Background}
\label{sec:background}

We begin by recalling the definitions of the Wasserstein distance and the Wasserstein barycenters. The Wasserstein distance, grounded in the OT theory, provides a principled framework for comparing probability distributions. In this work, we adopt the Kantorovich formulation of OT~\cite{peyre2019computational, montesuma2024recent}. Given empirical distributions $\hat{P}$ and $\hat{Q}$, represented by samples $x_i^{(P)} \sim P$ and $x_i^{(Q)} \sim Q$, we define:
\begin{equation*}
    \hat{Q}(x) = \frac{1}{n} \sum_{i=1}^{n} \delta(x - x_i^{(Q)})
\end{equation*}
and similarly for $\hat{P}(x)$. In this context $x_i^{(Q)}$ is call the support of $\hat{Q}(x)$. To compare distributions $P$ and $Q$ with $m$ and $n$ samples respectively, we consider the set of transport plans:
\begin{equation*}
    \Pi(P, Q) = \left\{ \pi \in \mathbb{R}_{+}^{n \times m} : \pi \mathbf{1}_m = \frac{1}{n} \mathbf{1}_n, \pi^T \mathbf{1}_n = \frac{1}{m} \mathbf{1}_m \right\}.
\end{equation*} 
Here, $\pi$ represents the plan for transporting mass from samples of $P$ to samples of $Q$. Given a cost matrix $C_{ij} = c(x_i^{(P)}, x_j^{(Q)})$, the Wasserstein distance is defined as
\begin{equation*}
    W_c(\hat{P}, \hat{Q}) = \min_{\pi \in \Pi(P, Q)} \sum_{i=1}^{n}\sum_{j=1}^{m}\pi_{ij}C_{ij},
\end{equation*}
where $c$ is a distance metric between samples. This Wasserstein distance facilitates the definition of a barycenter of distributions.

\begin{definition}
Given distributions \( P = \{P_k\}_{k=1}^{K} \) and weights \( \alpha \in \Delta^K \), the Wasserstein barycenter is defined as:
\begin{equation*}
    B^\star = B(\alpha; P) = \inf_{B} \sum_{k=1}^{K} \alpha_k W_c(P_k, B).
\end{equation*}
\end{definition}
\cite{Montesuma2023} introduce a novel framework for MSDA that combines Wasserstein barycenters with Dictionary Learning (DiL). In its classical form, DiL decomposes a set of vectors ${x_1, \ldots, x_N}$ into linear combinations of atoms ${p_1, \ldots, p_K}$, weighted by representation coefficients ${\alpha_1, \ldots, \alpha_N}$. The authors extend this principle to operate directly on empirical distributions.
\newline 
\\
Given datasets  $Q = \{\hat{Q}_{S_\ell}\}_{\ell=1}^{N_S} \cup \{\hat{Q}_T\}$
where $\{\hat{Q}_{S_\ell}\}_{\ell=1}^{N_S}$ denote the source domains and $\hat{Q}_T$ the target domain, the proposed Dataset Dictionary Learning (DaDiL) framework learns a set of atoms $P = \{\hat{P}_k\}_{k=1}^{K}
$
and barycentric coordinates 
$A = \{\alpha_\ell \in \Delta^K\}_{\ell=1}^{N}$
such that
\[
(P^\star, A^\star) = \arg\min_{P, A} \frac{1}{N} \sum_{\ell=1}^{N} f_\ell(\alpha_\ell, P).
\]

where $N = N_S + 1$, and $f_\ell$ is defined as
\begin{equation*}
    f_\ell(\alpha_\ell, P) = 
\begin{cases}
W_c(\hat{Q}_\ell, B(\alpha_\ell; P)) & \text{if } \hat{Q}_\ell \text{ is labeled}, \\
W_2(\hat{Q}_\ell, B(\alpha_\ell; P)) & \text{otherwise}.
\end{cases}
\end{equation*}

Here, \( W_c \) denotes a Wasserstein distance whose ground metric incorporates label information, while \( W_2 \) corresponds to the Wasserstein distance based on the squared Euclidean norm. The objective function is minimized with respect to the atom parameters \( (X(P_k), Y(P_k)) \) and the barycentric coordinates \( \alpha_\ell \) associated with each domain. DaDiL models each dataset distribution \( \hat{Q}_\ell \) as a labeled barycenter of the learned atoms \( \mathcal{P} = \{\hat{P}_k\}_{k=1}^{K} \). Once the dictionary \( (A, \mathcal{P}) \) is learned, the target domain can be reconstructed as a labeled barycenter \( B(\alpha_t; \mathcal{P}) \), from which labeled samples can be drawn to train a classifier for the target domain.

Eventually, \cite{clain2025decentralizedfederateddatasetdictionary} propose De-FedDaDiL, which extends DaDiL to a decentralized federated learning setting by distributing the optimization of $(\mathcal{P}, \mathcal{A})$ among multiple clients. Each client initializes its own version of the atoms $P_\ell^{(0)}$. At each iteration, a client randomly selects another client and sends it its current version of the atoms. Symmetrically, each client receives a version of the atoms from another client, aggregates the received version with its own, and then optimizes the aggregated atoms $P_\ell$ together with its barycentric coordinates $\alpha_\ell$ over $E$ local epochs. Only the atoms \(P_\ell\) are shared, while the barycentric coordinates \(\alpha_\ell\) remain private. This ensures privacy, since the original domain cannot be reconstructed without the barycentric coordinates.

\subsection{Background on GMM-DadiL}
\label{sec:proposed_approach_background}
\cite{GMMDADIL} propose a parametric extension of DaDiL, where atoms are represented as GMM rather than discrete distributions. Each atom $\mathcal{P}$ is parameterized as 

\begin{equation}
\Theta_{\mathcal{P}} = \{ (M(P_c), S(P_c), V(P_c)) \}_{c=1}^C    
\end{equation}
 
where \( C \) denotes the number of Gaussian components. For each component \( c \), \( M(P_c) \in \mathbb{R}^d \) represents its mean vector, \( S(P_c) \in \mathbb{R}^{d \times d} \) its diagonal covariance matrix, and \( V(P_c) \in \Delta^{n_{class}-1} \) its class assignment vector over the \( n_{class} \) possible classes. Similarly, each domain (or client) \( \mathcal{D} \) is modeled as a GMM parametrized as 
\begin{equation}
    \Theta_{\mathcal{D}} = \{ (M(D_c), S(D_c), V(D_c)) \}_{c=1}^{C_{\mathcal{D}}}
    \end{equation}
where $C_{\mathcal{D}}$
denotes the number of Gaussian components representing the domain distribution. 
For labeled source domains, a separate GMM is fitted for each class using a fixed number of components $n_{\text{comp\_per\_class}}$. 
The class-specific mixtures are then combined to form a single domain-level GMM with a total of $n_{\text{class}} \times n_{\text{comp\_per\_class}}$ components. 
For the unlabeled target domain, a single GMM with the same total number of components is fitted directly to the data.

Each domain is approximated by a Mixture-Wasserstein barycenter $B(\alpha, \mathcal{P})$. Learning the dictionary consists of estimating the atom parameters $\Theta_{\mathcal{P}}$ and barycentric coordinates $\alpha$ by solving

\begin{align}
(\alpha^\star, \Theta^\star_{\mathcal{P}})
&= \underset{\alpha, \Theta_{\mathcal{P}}}{\arg\min}
\Bigg[
    MW_2\big(Q_T, B(\alpha_T, \mathcal{P})\big)^2 \notag \\
&\quad + \sum_{\ell=1}^N
    SMW_2\big(Q_\ell, B(\alpha_\ell, \mathcal{P})\big)^2
\Bigg].
\end{align}

Where $MW_2(P,Q)$ is the Mixture-Wasserstein distance, which measures the Wasserstein distance between two GMM by optimally matching their components, and $SMW_2(P,Q)$ is its supervised extension that additionally penalizes mismatches between component labels in the transport plan. After learning $(\alpha^\star, \Theta_{\mathcal{P}}^\star)$, the target barycenter $B(\alpha_T, \mathcal{P})$ provides a labeled GMM, which can be used to generate labeled samples for classifier training.

A key motivation behind GMM-DaDiL is to reduce the computational cost of its empirical counterpart, DaDiL. By using a parametric Gaussian mixture model with diagonal covariances, GMM-DaDiL achieves strong adaptation performance with fewer parameters, making it lighter and more scalable while preserving the barycentric framework.

\section{Proposed Approach}
\label{sec:proposed_approach}

\subsection{Decentralized federated GMM-DaDiL}
We introduce DeFed-GMM-DaDiL, a fully decentralized extension of GMM-DaDiL, inspired by its non-parametric counterpart, De-FedDaDiL \cite{clain2025decentralizedfederateddatasetdictionary}.

\begin{figure}[]
    \centering
    \includegraphics[width=0.45\textwidth, trim={0.5cm 0.5cm 0.5cm 0.5cm}]{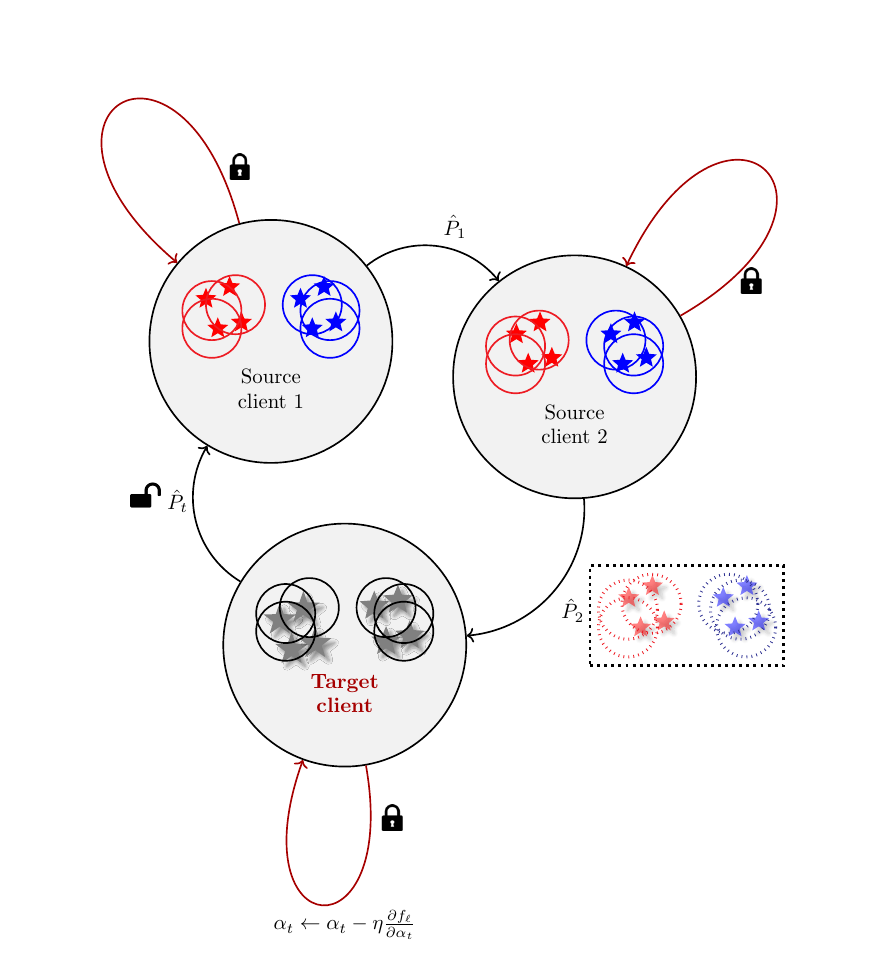}
\caption{\small DeFed-GMM-DaDiL: Clients initialize atoms and at each iterations receive atoms (\includegraphics[width=0.9em]{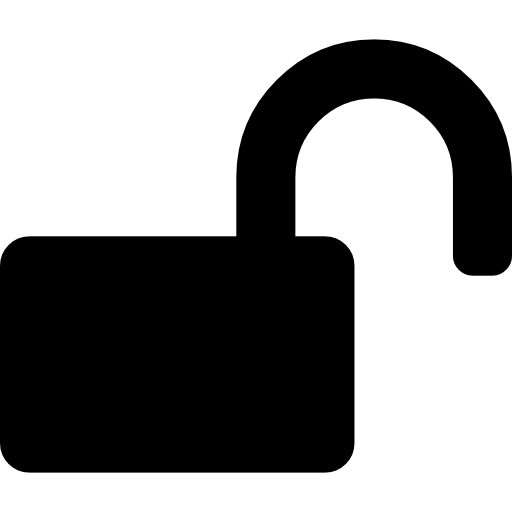}$\mathcal{P}$) from random peer(s). Clients update their models based on their own and/or received version, while keeping barycentric coordinates private (\includegraphics[width=0.9em]{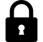}$\alpha$). Stars: domain datasets; circles: GMM components (three per domain); dotted circles: learned atom GMM components.}
    \label{fig:De_FedDaDil}
\end{figure}

In DeFed-GMM-DaDiL, both the initialization and optimization of the atoms $(\mathcal{P})$ and barycentric coordinates $(\mathcal{A})$ are fully distributed across clients, enabling completely decentralized operation. DeFed-GMM-DaDiL is illustrated in Figure \ref{fig:De_FedDaDil}. In practice, each client represents its local dataset as a GMM, following the procedure described in Section \ref{sec:proposed_approach_background}. It then initializes a local dictionary \(\{P_\ell^{(0)}, \alpha_\ell^{(0)}\}\), where \(P_\ell\) denotes the GMM atoms and \(\alpha_\ell\) the barycentric coordinates.

At each round, clients exchange atom parameters using one of two strategies:

\begin{itemize}
    \item \textbf{Replacement:} each client randomly selects one peer and replaces its local atoms with those received.
    \item \textbf{Aggregation:} each client randomly selects two peers and performs barycentric averaging, computing the weighted Wasserstein mean of the two received and its own atoms. Using two peers rather than one reduces instability and improves the smoothness of updates, as single-peer averaging can result in abrupt updates before local re-optimisation.
\end{itemize}

Only the atoms are shared; barycentric coordinates remain private. Each client computes a loss: labeled sources use supervised $\text{SMW}_2$, and the unlabeled target uses unsupervised $\text{MW}_2$, as in GMM-DaDiL. Both atoms and barycentric coordinates are updated via gradient descent. After the iterations, each client has its optimized atoms and barycentric coordinates $(P_\ell^*, \alpha_\ell^*)$. We detail the strategy in Algorithm \ref{algo:defed_gmm_dadil} where each client $\ell$ holds atoms $P_\ell = \{\mathcal{P}_k\}_{k=1}^K$, with each atom $\mathcal{P}_k$ parameterized as $\Theta_{\mathcal{P}_k} = \{(M(P_c), S(P_c), V(P_c))\}_{c=1}^C$ 

\begin{algorithm}[]
\caption{\small DeFed-GMM-DaDiL}
\label{algo:defed_gmm_dadil}
\small 
\KwIn{\mbox{Sources $\{Q_{S_\ell}\}_{\ell=1}^N$, target $Q_T$, iterations $N_{it}$}}

\KwOut{Local parameters $\{P_\ell, \alpha_\ell\}_{\ell=1}^N$}

\BlankLine
\textbf{Init (each client $\ell$):} 
$M(P_\ell) \sim \mathcal{N}(0,I)$, 
\mbox{$S(P_\ell) \gets 1$, $V(P_\ell) \gets \mathbf{1}_{n_{class}}/n_{class}$, $\alpha_\ell \gets 1/K$}

\For{$t=1,\dots,N_{it}$}{
    \If{Replacement}{
        Client $\ell$ selects one random peer $r \neq \ell$; 
        $P_\ell \gets P_r$ 
    }
    \If{Aggregation}{
         Client $\ell$ selects random peers $r_1,r_2 \neq \ell$; 
        $P_\ell \gets \text{Average}(P_\ell,P_{r_1},P_{r_2})$ 
    }

    \If{source data}{
        $L_\ell \gets L_\ell + \text{SMW}_2^2(Q_{S_\ell}, B(\alpha_\ell,P_\ell))$;
    }
    \If{target data}{
        $L_\ell \gets L_\ell + \text{MW}_2^2(Q_T, B(\alpha_\ell,P_\ell))$;
    }

    $M(P_\ell) \gets M(P_\ell) - \eta \nabla_{M(P_\ell)} L_\ell$; \\
    $S(P_\ell) \gets \text{Proj}_{\mathbb{R}_+^d}\big(S(P_\ell) - \eta \nabla_{S(P_\ell)} L_\ell\big)$; \\
    $V(P_\ell) \gets V(P_\ell) - \eta \nabla_{V(P_\ell)} L_\ell$; \\
    $\alpha_\ell \gets \text{Proj}_\Delta\big(\alpha_\ell - \eta \nabla_{\alpha_\ell} L_\ell\big)$;
}
\Return{For each client $\ell$: $(P_\ell, \alpha_\ell)$}
\end{algorithm}

\textbf{Complexity:} We compare the computational and communication costs of non-parametric De-FedDaDiL (introduced in Section \ref{sec:background}) and parametric DeFed-GMM-DaDiL.  
In De-FedDaDiL, the dominant local computation per client is solving OT on mini-batches of data. For a batch size $n_b$, $M$ mini-batches per iteration, $A$ atoms, and $N_{\text{it}}$ local iterations, the cost scales as $O(N_{\text{it}} \cdot M \cdot A \cdot n_b^3 \log n_b)$. Communication requires exchanging the empirical atoms, with cost $O(A \cdot d)$, where $d$ is the data dimensionality.
In DeFed-GMM-DaDiL, each atom is represented as a Gaussian mixture with $n_{c}$ classes and $C$ components per class, with diagonal covariance matrices. 
The local computation per iteration scales as $O(N_{\text{it}} \cdot A \cdot n_{c} \cdot C^3 \log C)$, which is typically smaller in practice than in De-FedDaDiL because $C$ is usually much smaller than the number of points $n_b$ used in empirical atoms. 
Communication consists of exchanging the GMM parameters (means, diagonal covariances, and mixture weights) for each atom. 
Although the big-O complexity with respect to the number of atoms $A$ and the data dimension $d$ remains $O(A \cdot d)$, the constant is significantly reduced since $d_{\text{GMM}} \ll d \cdot n_b$.

\subsection{Empirical results}
\begin{table*}[t]
    \centering
    \footnotesize
    \setlength{\tabcolsep}{1pt} 
    \renewcommand{\arraystretch}{0.7}
    
    \begin{minipage}{0.31\linewidth}
        \centering
        \begin{tabular}{@{}lcccc>{\columncolor[gray]{0.9}}c@{}}
            \toprule
            \textbf{Method} & \textbf{C} & \textbf{B} & \textbf{I} & \textbf{P} & \textbf{Avg$\uparrow$} \\
            \midrule
            FedAVG & 96.7 & 65.8 & 94.2 & 77.5 & 83.6\\
            FedProx & 96.7 & 65.8 & 93.3 & 76.7 & 83.1\\
            \midrule
            $f$-DANN & 96.7 & 64.2 & 87.5 & 80.0 & 82.1 \\
            $f$-WDGRL & 92.5 & 63.3 & 86.7 & 74.2 & 79.2 \\
            FADA & 95.0 & 64.2 & 90.0 & 74.2 & 80.9 \\
            KD3A & 93.3 & 69.2 & 95.5 & 73.3 & 82.8 \\
            Co-MDA & 94.2 & 65.0 & 91.5 & 78.0 & 82.2 \\
            \midrule
            FedDaDiL-E & 98.3 & 69.2 & 93.3 & 81.6 & 85.6\\
            FedDaDiL-R & 98.3 & 69.2 & 95.0 & 80.0 & 85.6\\
            \midrule
            De-FedDaDiL-E & 98.3 & 70.8 & 98.3 & 79.2 & 86.6 \\
            De-FedDaDiL-R & 98.3 & 70.8 & 98.3 & 79.2 & 86.6 \\
            \midrule
            DeFed-GMM & 97.5 & 69.2 & 96.7 & 77.5 & 85.2 \\
            \bottomrule
        \end{tabular}
        \\
        \footnotesize\textbf{(a) ImageCLEF}
    \end{minipage}\hfill
    \begin{minipage}{0.31\linewidth}
        \centering
        \begin{tabular}{@{}lccc>{\columncolor[gray]{0.9}}c@{}}
            \toprule
            \textbf{Method} & \textbf{A} & \textbf{d} & \textbf{W} & \textbf{Avg$\uparrow$} \\
            \midrule
            FedAVG & 67.5 & 95.0 & 96.8 & 86.4\\
            FedProx & 67.4 & 96.0 & 96.8 & 86.7\\
            \midrule
            $f$-DANN & 67.7 & 99.0 & 95.6 & 87.4 \\
            $f$-WDGRL & 64.8 & 99.0 & 94.9 & 86.2 \\
            FADA & 62.5 & 97.0 & 93.7 & 84.4 \\
            KD3A & 65.2 & 100.0 & 98.7 & 88.0 \\
            Co-MDA & 64.8 & 99.8 & 98.7 & 87.8\\
            \midrule
            FedDaDiL-E & 71.2 & 100.0 & 98.2 & 89.8 \\
            FedDaDiL-R & 70.6 & 100.0 & 99.4 & 90.0 \\
            \midrule
            De-FedDaDiL-E & 68.3 & 99.7 & 98.7 & 88.9 \\
            De-FedDaDiL-R & 67.9 & 99.0 & 99.4 & 88.8 \\
            \midrule
            DeFed-GMM & 69.1 & 99.0 & 98.1 & 88.8 \\
            \bottomrule
        \end{tabular}
        \\
        \footnotesize\textbf{(b) Office 31}
    \end{minipage}\hfill
    \begin{minipage}{0.31\linewidth}
        \centering
        \begin{tabular}{@{}lcccc>{\columncolor[gray]{0.9}}c@{}}
            \toprule
            \textbf{Method} & \textbf{A} & \textbf{C} & \textbf{P} & \textbf{R} & \textbf{Avg$\uparrow$} \\
            \midrule
            FedAVG & 72.9 & 62.2 & 83.7 & 85.0 & 76.0\\
            FedProx & 70.8 & 63.7 & 83.6 & 83.1 & 75.3\\
            \midrule
            $f$-DANN & 70.2 & 65.1 & 84.8 & 84.0 & 76.0 \\
            $f$-WDGRL & 68.2 & 64.1 & 81.3 & 82.5 & 74.0 \\
            FADA & -- & -- & -- & -- & -- \\
            KD3A & 73.8 & 63.1 & 84.3 & 83.5 & 76.2 \\
            Co-MDA$^{\star}$ & 74.4 & 64.0 & 85.3 & 83.9 & 76.9\\
            \midrule
            FedDaDiL-E & 75.7 & 64.7 & 85.9 & 85.6 & 78.0 \\
            FedDaDiL-R & 76.5 & 65.2 & 85.9 & 84.2 & 78.0 \\
            \midrule
            De-FedDaDiL-E & 76.3 & 63.2 & 84.6 & 85.1 & 77.3 \\
            De-FedDaDiL-R & 76.1 & 63.5 & 84.1 & 84.9 & 77.1 \\
            \midrule
            DeFed-GMM & 77.0 & 63.9 & 85.1 & 85.4 & 77.9 \\
            \bottomrule
        \end{tabular}
        \\
        \footnotesize\textbf{(c) Office-Home}
    \end{minipage}
    
    {\footnotesize\textbf{Abbreviations:} 
    (a) C:Caltech, B:Bing, I:ImageNet, P:Pascal;
    (b) A:Amazon, d:dSLR, W:Webcam; 
    (c) A:Art, C:Clipart, P:Product, R:Real-World}

        \caption{\small Experimental Results on decentralized MSDA benchmarks. $\uparrow$ denotes that higher is better.}
            \label{tab:fed_da_results}
\end{table*}

We compare DeFed-GMM-DaDiL with other decentralized MSDA approaches. The results are summarized in Table \ref{tab:fed_da_results}. Experiments are conducted on ImageCLEF~\cite{caputo2014imageclef}, Office-31~\cite{saenko2010adapting}, and Office-Home~\cite{venkateswara2017deep}. We include standard federated baselines FedAvg~\cite{mcmahan2017communication} and FedProx~\cite{li2020federated} and three state-of-the-art decentralized MSDA methods: FADA~\cite{peng2019federated}, KD3A~\cite{feng2021kd3a}, and Co-MDA~\cite{liu2023co}. We further compare DeFed-GMM-DaDiL with its non-parametric variants — De-FedDaDiL \cite{clain2025decentralizedfederateddatasetdictionary}, which operates in a fully decentralized federated setting, and FedDaDiL \cite{espinoza2024federated}, its federated counterpart. For completeness, we also evaluate adapted federated versions of classical domain adaptation methods, $f$-DANN~\cite{ganin2016domain,peng2019federated} and $f$-WDGRL~\cite{shen2018wasserstein}.

DeFed-GMM-DaDiL achieves strong performance across all benchmarks, outperforming standard federated baselines FedAVG and FedProx, as well as adversarial and distillation-based methods including $f$-DANN, $f$-WDGRL, FADA, and KD3A.
Beyond accuracy, DeFed-GMM-DaDiL remains considerably lighter thanks to its compact parametric GMM representation, which reduces the number of trainable parameters compared to its non-parametric counterpart De-FedDaDiL. Furthermore, unlike most competing methods, DeFed-GMM-DaDiL, along with De-FedDaDiL, operates in a fully decentralized setting without relying on a central server. Overall, these characteristics highlight its combination of high predictive performance and model compactness within a fully decentralized setting.

\paragraph{Analyzing Consensus Among Clients’ Atoms:}

In the DeFed-GMM-DaDiL framework, each client maintains locally optimized atoms with corresponding barycentric coordinates. Although initializations vary across clients, iterative sharing (via averaging or replacement) should progressively align the atoms. To assess consensus,
at each iteration $t$, we compute for each client, a set of barycenters using a common set of weight combinations
\begin{equation}
\mathcal{W}_t = \{w^{(1)}, \dots, w^{(L)}\}.
\end{equation}
which provides a discrete approximation of the hull of possible barycenters at that iteration. For a given weight \(w^{(\ell)}\) at iteration \(t\), the pairwise discrepancy between clients \(i\) and \(j\) is measured using the Wasserstein-2 distance:

\begin{equation}
D_{ij}^{(\ell)}(t) = W_2\!\big(\mu_i^{(w^{(\ell)})}(t), \mu_j^{(w^{(\ell)})}(t)\big),
\end{equation}
and their worst-case discrepancy for iteration $t$ is defined as:
\begin{equation}
M_{ij}(t) = \max_{\ell} D_{ij}^{(\ell)}(t).
\end{equation}

We track the set of maximal distances between all client pairs,
$\{ M_{ij}(t) \mid 1 \le i < j \le N \}$
over time to provide a quantitative measure of global consensus, with \(N\) being the total number of clients.

\begin{figure}[t]
    \centering
    \includegraphics[width= 1\linewidth]{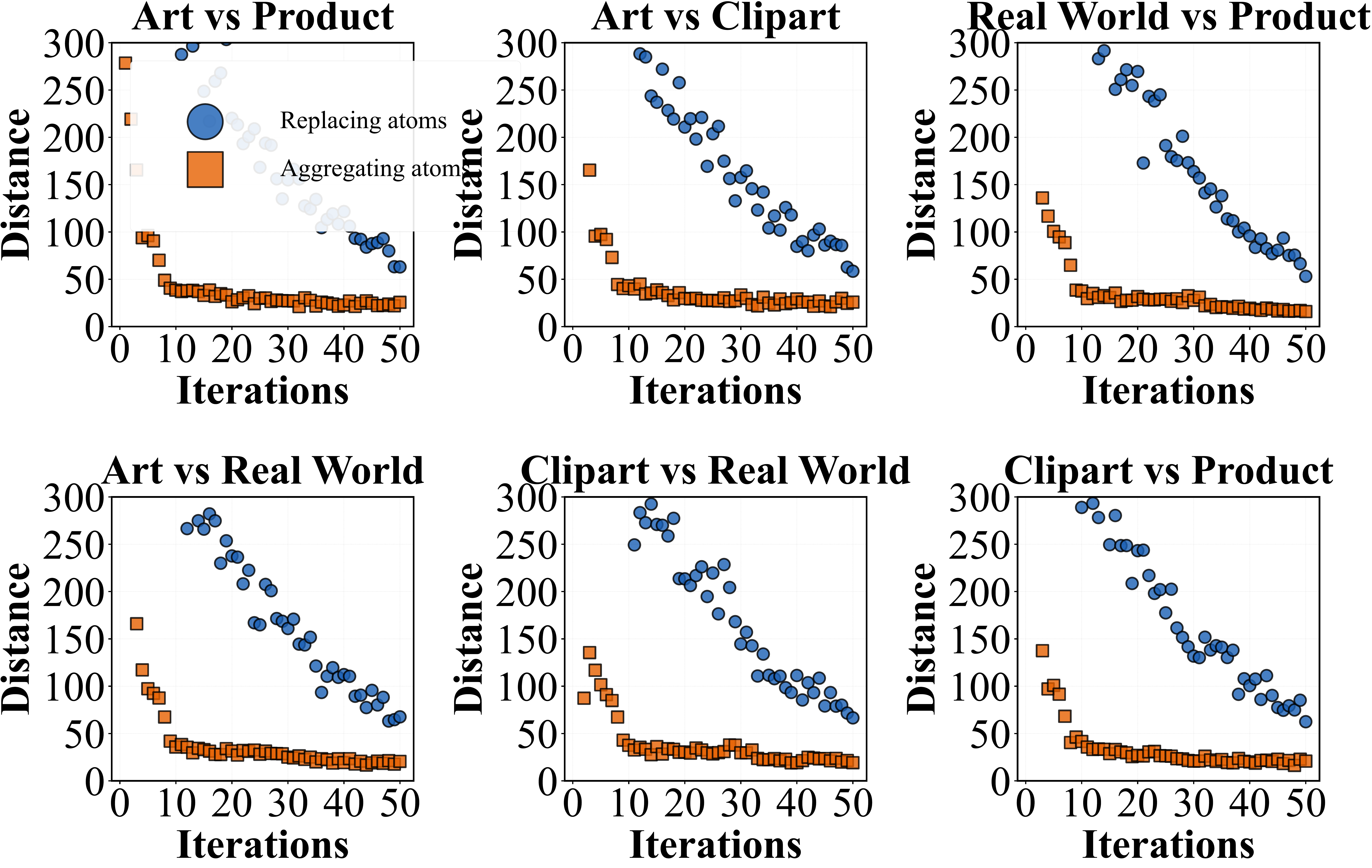}
    \caption{\small Maximum discrepancy $M_{ij}$ between domain barycenters over training iterations, for replacement and aggregation strategies. Target domain is Real World.}
    \label{fig:replacing_vs_aggregated_consensus}
\end{figure}

\begin{figure}
    \centering
    \includegraphics[width=0.9\linewidth, trim={2cm 3cm 2cm 3cm}, clip]{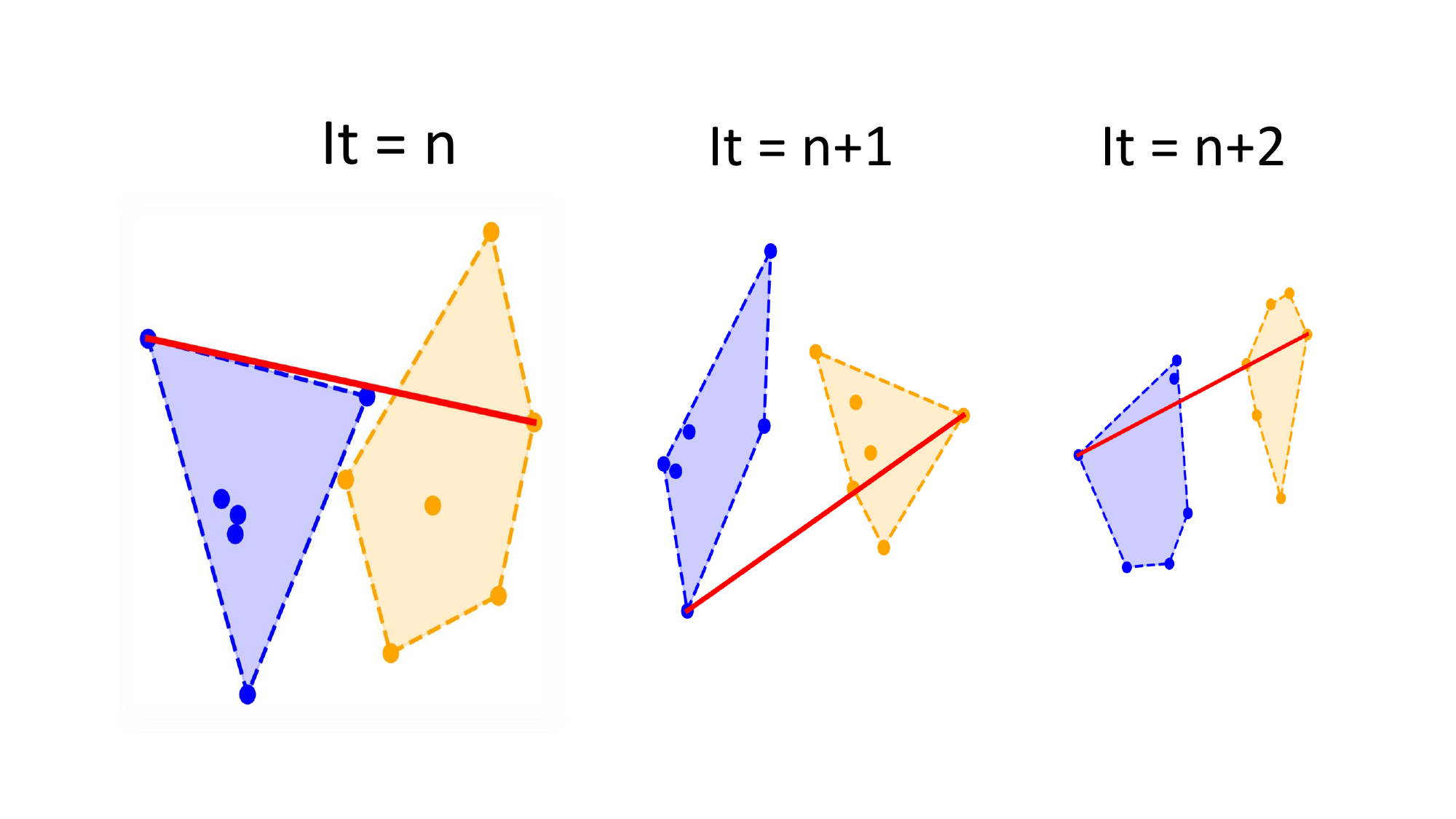}
\caption{\small Evolution of consensus between two clients. Each panel shows the hulls of barycenters for two clients (blue and yellow) at iterations $t$, $t+1$, and $t+2$. The red segment highlights the maximal cross-client Wasserstein-2 distance, $M_{ij}(t)$, which progressively decreases over iterations, indicating increasing alignment of the clients’ local atoms.}
   \label{fig:illustration_bary}
\end{figure}

A decrease in these distances reflects the progressive alignment of client atoms, providing evidence of consensus. Figure \ref{fig:illustration_bary} provides a schematic illustration of barycenter consensus, showing how the hull contracts as domains align over iterations.
Figure \ref{fig:replacing_vs_aggregated_consensus} presents empirical results on the OfficeHome benchmark with Real World as the target domain. It shows the evolution of \( M_{ij} \), the maximum discrepancy between domain barycenters, over the training iterations.
  It compares the two distinct strategies for atom sharing across clients: the replacement approach (depicted in blue), where clients completely substitute their local atoms with received versions at each iteration, and the aggregation strategy (shown in orange), where clients aggregate received atoms with their local versions. Both strategies demonstrate convergence behavior, though at differing rates. The replacement approach exhibits slower convergence compared to the aggregation strategy. Convergence in the replacement setting is still achieved due to three key factors: frequent atom circulation among a limited client pool; initialization from the same statistical process (GMM from a normal distribution), providing structurally similar starting models; and inherent similarities in the data domains. These gradually align the atoms. However, aggregation achieves higher similarity because it employs the same mechanisms, enhanced by the smoothing effect of averaging, which reduces—though does not eliminate—client-specific variance.

\section{Missing Classes: implications for target accuracy, reconstruction, and consensus}
\label{sec:missing_classes}
\begin{figure*}[t]
    \centering

    \begin{subfigure}[t]{0.48\linewidth}
        \centering
        \includegraphics[width=\linewidth]{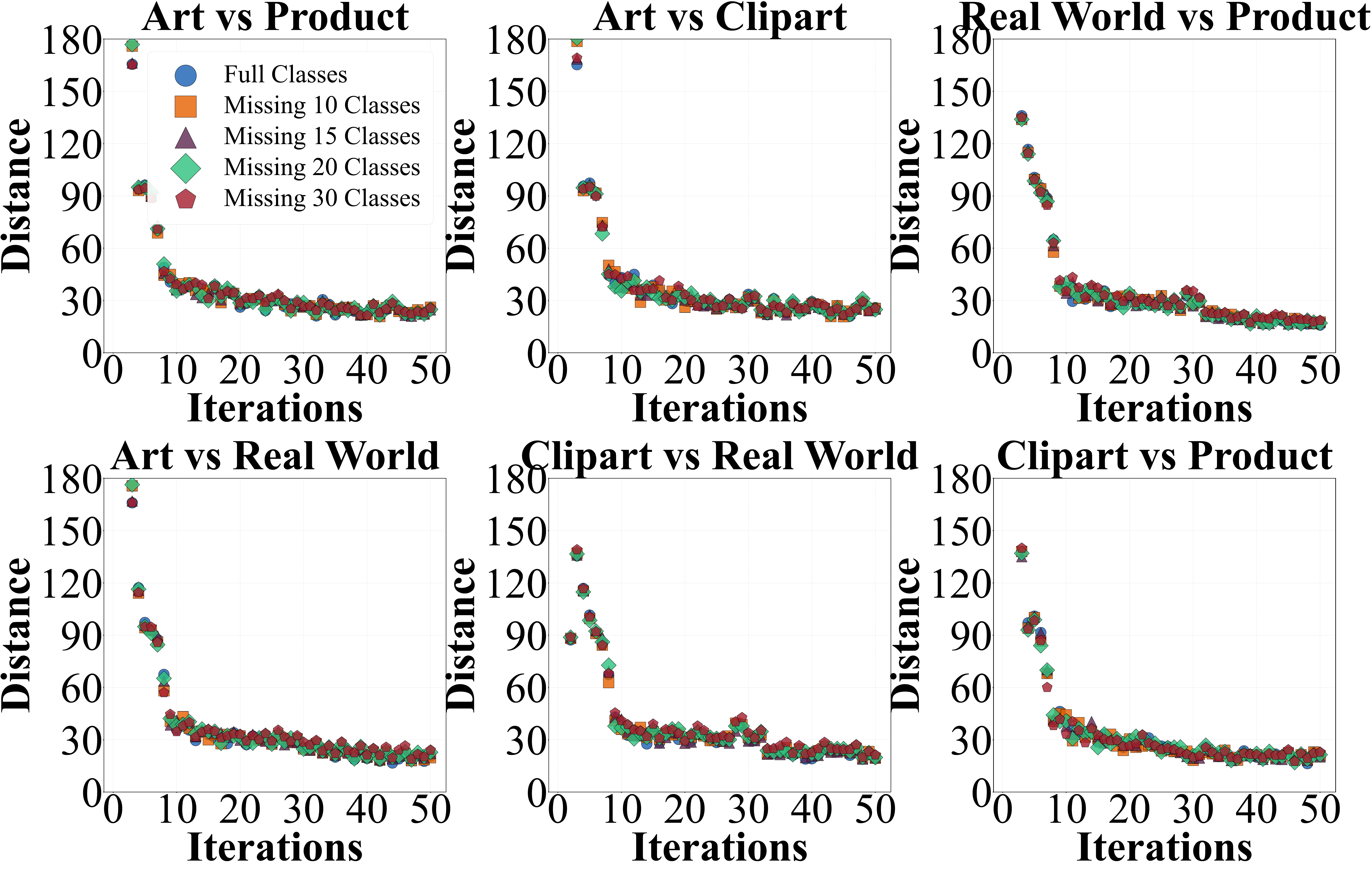}
        \caption{Aggregation strategy}
        \label{fig:aggregated}
    \end{subfigure}
    \hfill
    \begin{subfigure}[t]{0.48\linewidth}
        \centering
        \includegraphics[width=\linewidth]{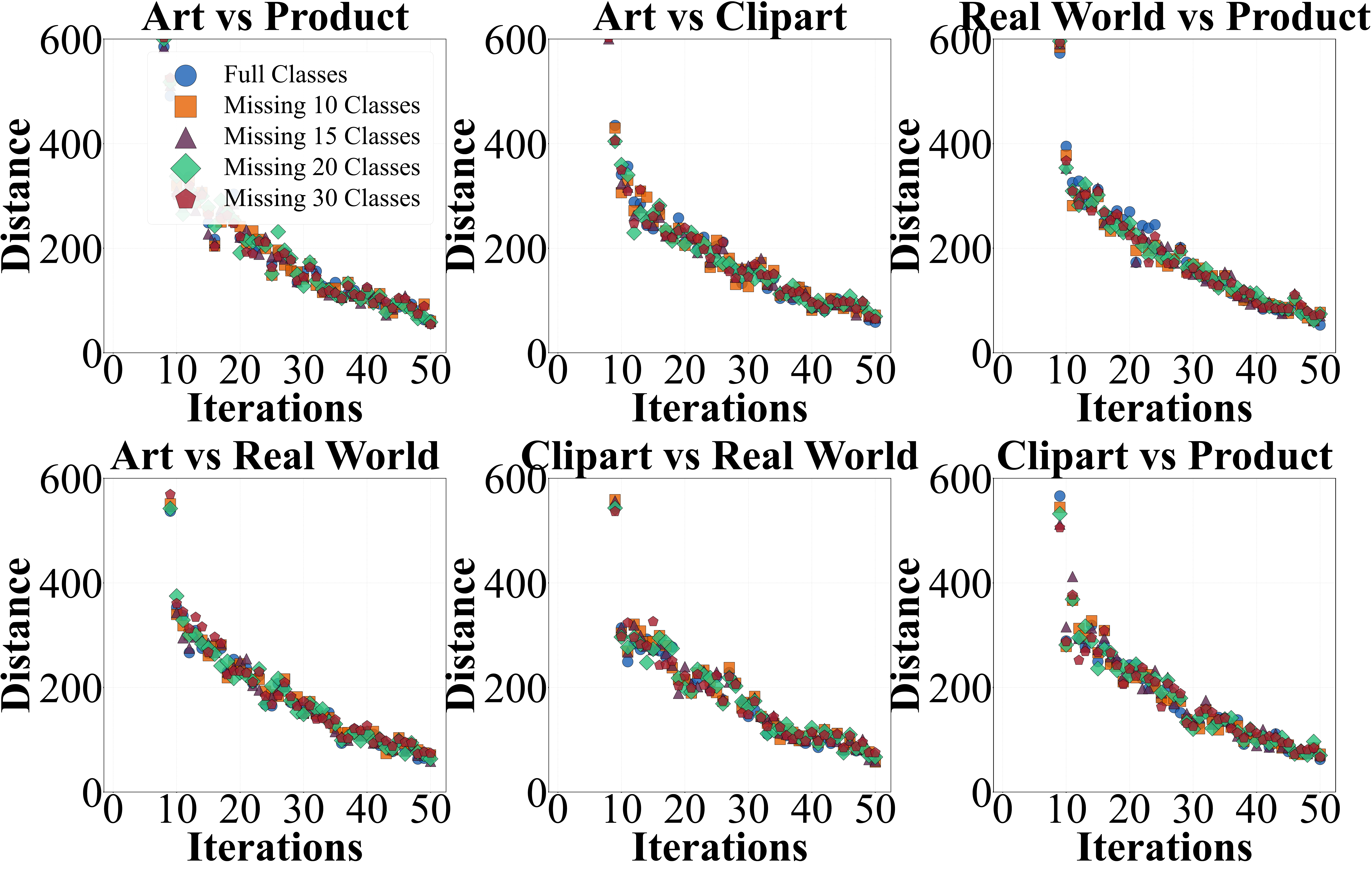}
        \caption{Replacement strategy}
        \label{fig:non_aggregated}
    \end{subfigure}

\caption{Maximum discrepancy, $M_{ij}$, between domain barycenters, over training iterations for replacement and aggregation strategies under different levels of missing classes. The target domain is the Real World.}
\label{fig:aggreg_vs_nonaggreg}
\end{figure*}


 In this section, we analyze the stability of our approach under varying fractions of target missing classes. 
We begin by examining target accuracy on benchmark datasets and visualizing the reconstruction of missing classes with t-SNE visualization. 
We then investigate the stability of the target barycentric envelope and conclude with an analysis of consensus stability across clients.


\subsection{Accuracy and reconstruction visualization}

\begin{figure}[t]
    \centering
    \includegraphics[width=\linewidth]{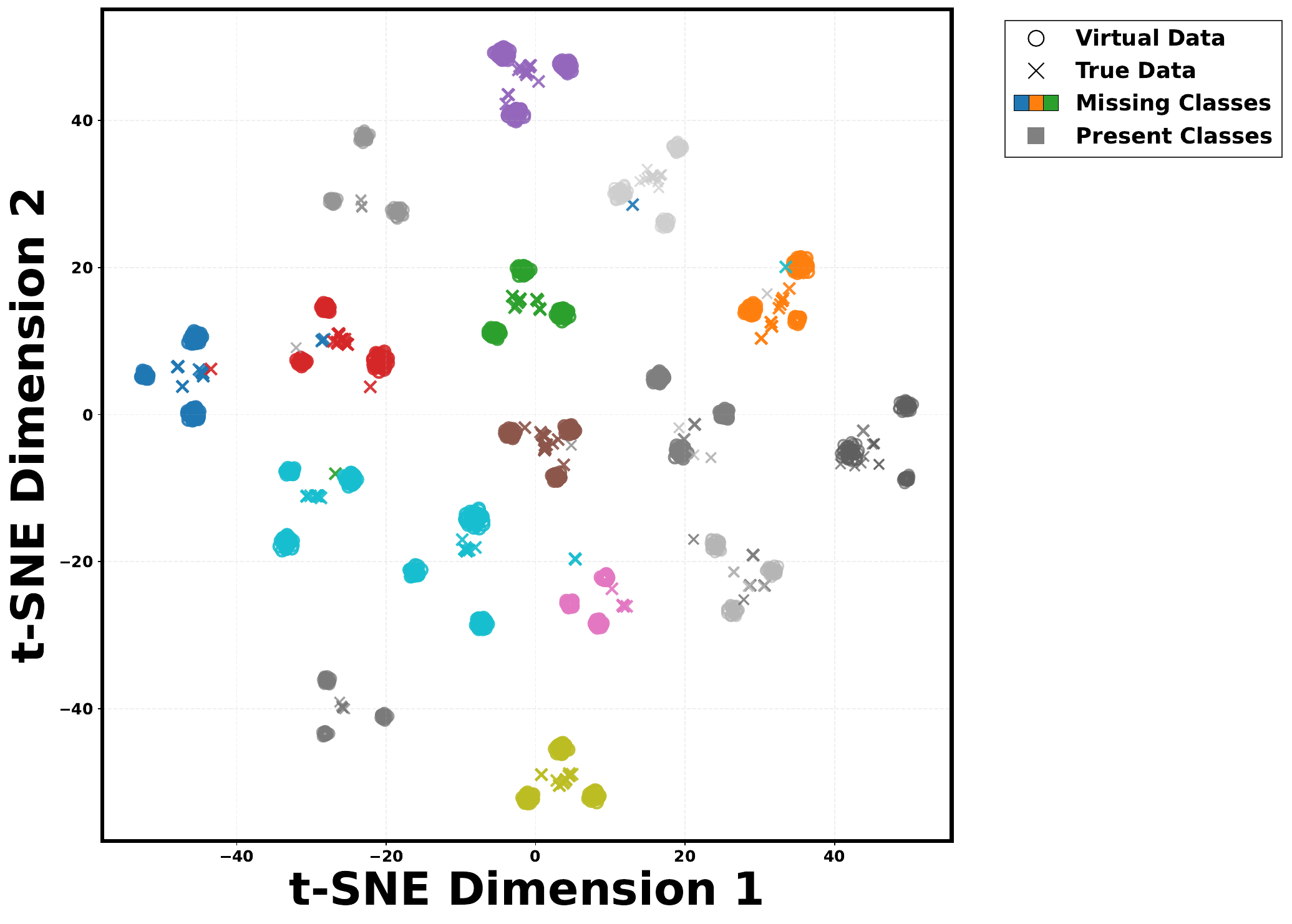}
\caption{t-SNE visualization of learned target atoms under missing-class conditions. Present classes are shown in grayscale, and 10 randomly selected missing classes (out of 30) are highlighted to illustrate how missing classes are reconstructed in the target domain.}

    \label{fig:tsne}
\end{figure}

We first evaluate the effect of missing classes on target accuracy across the three benchmark datasets (ImageCLEF, Office-31, and Office-Home). Table \ref{tab:accuracy_missing} reports classification performance under increasing fractions of missing classes, showing how accuracy evolves as target training data becomes incomplete. In these experiments, we run DeFed-GMM-DaDiL with different percentages of missing classes in the target training dataset, while evaluation is carried out on a separate target test set containing the full set of classes. We assume that the target client knows the total number of classes, even if some are absent from its training data. Across datasets, the average target accuracy decreases moderately but remains high as the fraction of missing classes increases, with mean performance above 82\% on ImageCLEF, 85\% on Office-31, and 76\% on Office-Home. This demonstrates the robustness of the DeFed-GMM-DaDiL framework, which maintains stable performance despite missing target classes. To complement these results, we provide a visualization using t-SNE, a dimensionality reduction technique that projects high-dimensional data into two dimensions while preserving local neighborhood structure. Specifically, once the target atoms are learned through federated training, we compute their barycenter (the target barycenter) and generate virtual samples from it. We then apply t-SNE jointly to the real and virtual target datasets, enabling a direct visual comparison between reconstructed and real samples. Figure \ref{fig:tsne} presents the t-SNE visualization, showing ten randomly selected present classes in grayscale and ten missing classes (out of thirty). The figure illustrates that even when the target domain has missing classes, the federated barycenter effectively reconstructs them.

\begin{table*}[t]
    \centering
    \footnotesize
    \setlength{\tabcolsep}{1.8pt} 
    \renewcommand{\arraystretch}{0.95} 
    
    \begin{minipage}{0.32\linewidth}
        \centering
        \begin{tabular}{@{}lcccc>{\columncolor[gray]{0.9}}c@{}}
            \toprule
            \textbf{\scriptsize \shortstack[l]{Removed\\Samples (\%)}} & \textbf{C} & \textbf{B} & \textbf{I} & \textbf{P} & \textbf{Avg} \\
            \midrule
            0\% & 97.5 & 69.16 & 96.66 & 77.5 & 85.2 \\
            2-16\% & 96.66 & 67.5 & 92.5 & 73.33 & 82.5 \\
            3-25\% & 95.83 & 68.33 & 92.5 & 74.16 & 82.7 \\
            4-33\% & 96.66 & 68.33 & 95.00 & 70.83 & 82.7 \\
            6-50\% & 95.83 & 65.83 & 92.5 & 75.00 & 82.29 \\
            \bottomrule
        \end{tabular}
        \\[2pt]
        \footnotesize\textbf{(a) ImageCLEF}
    \end{minipage}\hfill
    \begin{minipage}{0.32\linewidth}
        \centering
        \begin{tabular}{@{}lccc>{\columncolor[gray]{0.9}}c@{}}
            \toprule
            \textbf{\scriptsize \shortstack[l]{Removed\\Samples (\%)}} & \textbf{A} & \textbf{d} & \textbf{W} & \textbf{Avg} \\
            \midrule
            0\% & 69.14 & 99.0 & 98.14 & 88.75 \\
            5-16\% & 65.60 & 99.0 & 97.48 & 87.36 \\
            7-22\% & 66.84 & 99.0 & 96.22 & 87.35 \\
            10-32\% & 62.41 & 98.0 & 97.48 & 85.96 \\
            14-45\% & 65.07 & 97.0 & 96.22 & 86.09 \\
            \bottomrule
        \end{tabular}
        \\[2pt]
        \footnotesize\textbf{(b) Office-31}
    \end{minipage}\hfill
    \begin{minipage}{0.32\linewidth}
        \centering
        \begin{tabular}{@{}lcccc>{\columncolor[gray]{0.9}}c@{}}
            \toprule
            \textbf{\scriptsize\shortstack[l]{Removed\\Samples (\%)}} & \textbf{A} & \textbf{C} & \textbf{P} & \textbf{R} & \textbf{Avg} \\
            \midrule
            0\% & 76.95 & 63.91 & 85.13 & 85.43 & 77.85 \\
            10-15\% & 74.89 & 63.11 & 82.65 & 85.83 & 76.12 \\
            15-23\% & 74.89 & 62.31 & 83.55 & 84.28 & 76.26 \\
            20-30\% & 75.51 & 62.65 & 83.78 & 84.05 & 76.50 \\
            30-46\% & 75.30 & 62.88 & 81.98 & 84.51 & 76.17 \\
            \bottomrule
        \end{tabular}
        \\[2pt]
        \footnotesize\textbf{(c) Office-Home}
    \end{minipage}
    
    \vspace{3pt}
{\footnotesize
\textbf{Abbreviations:}  
(a) C: Caltech, B: Bing, I: ImageNet, P: Pascal;  
(b) A: Amazon, d: dSLR, W: Webcam;  
(c) A: Art, C: Clipart, P: Product, R: Real-World.  

\textbf{Note:} First column shows fraction of target classes removed; percentages cover low to high incompleteness.
}

\caption{Target accuracy for varying fractions of missing classes.}

    \label{tab:accuracy_missing}
\end{table*}

\subsection{Barycentric envelope stability analysis}

This subsection investigates the stability of the target barycentric envelope on the Office-Home dataset (65 classes) when several target classes are missing. We evaluate whether the target barycentric weights remain consistent under missing-class scenarios, to determine if the barycentric representation can capture the full target distribution from incomplete data.
The DeFed-GMM-DaDiL algorithm is first trained using only the source clients. 
Then, a dictionary of atoms is selected from a random source client and kept fixed. 
Let $P^\star = \{ P_k^\star \}_{k=1}^K$ denote the atoms from this dictionary. 
We then optimize the target barycentric weights for the fixed atoms $P^\star$, minimizing the $MW^2$ distance between the target domain and its barycentric reconstruction: $\min_{\alpha_t \in \mathbb{R}^K} \mathrm{MW}^2\big(Q_T, B(\alpha_t; P^\star)\big)$
where $B(\alpha_t; P^\star)$ denotes the barycenter associated with the fixed atoms $P^\star$ and the barycentric weights $\alpha_t$. 
We consider two scenarios:
\begin{itemize}
    \item \small \textbf{Complete Target} ($Q_T^{\mathrm{full}}$): all 65 classes are present.
    \item \small \textbf{Reduced Target} ($Q_T^{\mathrm{red}}$): 10, 20, or 30 classes are removed.
\end{itemize}

The weights are updated via gradient descent, followed by a Euclidean projection onto the probability simplex after the final iteration:
$\alpha_t^\star = \Pi_{\Delta_K}\left(\alpha_t^{(N)}\right)$

We assess the stability of the barycentric envelope by comparing the similarity between the final weight vectors $\alpha_t^{\mathrm{full}}$ and $\alpha_t^{\mathrm{red}}$, the relative reconstruction error $\mathrm{MW}^2(Q_T^{\mathrm{red}}, B(\alpha_t^{\mathrm{red}}; P^\star))$ versus $\mathrm{MW}^2(Q_T^{\mathrm{full}}, B(\alpha_t^{\mathrm{full}}; P^\star))$, and the classification accuracy of virtual samples drawn from the final barycenters. Table \ref{tab:bary_weights} summarizes the final barycentric weights (the number of atoms $K$ is 3) and reconstruction performance across the different missing-class configurations.

\begin{table}[t]
\centering
\resizebox{\linewidth}{!}{%
\begin{tabular}{lccc|cc}
\toprule
\textbf{Missing Classes} & $\alpha_1$ & $\alpha_2$ & $\alpha_3$ & \textbf{MW$^2$ Loss} & \textbf{Accuracy (\%)} \\
\midrule
NA          & 0.0000 & 0.4812 & 0.5188 & 944.45  & 82.65 \\
10 (15\%)   & 0.0000 & 0.4822 & 0.5178 & 1041.00 & 81.86 \\
20 (30\%)   & 0.0226 & 0.4760 & 0.5014 & 1156.00 & 83.10 \\
30 (46\%)   & 0.0455 & 0.4690 & 0.4855 & 1275.00 & 82.77 \\
\bottomrule
\end{tabular}}
\caption{Final barycentric weights ($K=3$) and reconstruction performance across missing-class configurations.}
\label{tab:bary_weights}
\end{table}

The MW$^2$ reconstruction loss increases monotonically as the fraction of missing classes grows, indicating that the barycenter becomes progressively harder to match when information is absent in the target domain. Nevertheless, the overall increase remains moderate, suggesting that the learned atoms retain strong generalization ability. Classification accuracy remains stable across all configurations, fluctuating within a narrow band of 81.9--83.1\%. This indicates that the barycentric reconstruction preserves the discriminative structure of the data even under significant class removal. The barycentric weight vectors further confirm this stability: for 10 (15\%) missing classes, the solution is nearly indistinguishable from the complete-target case. At 20~(30\%) and 30~(46\%) missing classes, the weights adjust gradually, redistributing mass while maintaining overall proportions. This shows the robustness of the barycentric envelope, which adapts smoothly without collapse. Overall, these results suggest that the barycentric representation captures the underlying target distribution even when part of the target data is missing.

\subsection{Consensus stability}

Beyond evaluating the barycentric envelope, we also examine the consensus between clients’ atoms when the target domain has missing classes. The objective is to assess whether decentralized updates still drive local dictionaries toward a common representation, even when the target distribution is incomplete. To this end, we monitor the maximum Wasserstein distance, $M_{ij}$, between client barycenters across iterations under different missing-class configurations for both strategies. This provides a complementary view of stability, focusing on the alignment of client atoms rather than the quality of target reconstruction. The experiments are conducted on the Office-Home dataset and the results are illustrated in Figure \ref{fig:aggreg_vs_nonaggreg}.
The trajectories show that the distances between domain barycenters decrease steadily across iterations in both scenarios, confirming that consensus emerges consistently. Convergence rates are comparable between different levels of missing classes, and the curves stabilize at nearly identical values. This indicates that clients achieve a similar degree of alignment even under substantial class removal. Overall, these results confirm the stability of our approach and its ability to recover missing classes despite incomplete target information.

\section{Conclusion}
\label{sec:conclusion}

This paper presents the DeFed-GMM-DaDiL framework, a federated and fully decentralized adaptation of the GMM-DaDiL framework. Its decentralized design allows multiple clients to collaboratively adapt to a target domain efficiently, without relying on a central server. Experiments show that it achieves competitive performance compared to existing benchmarks, remains robust when target domains have missing classes, and drives clients toward consensus even under partial data coverage.



\newpage

\bibliographystyle{named}
\bibliography{ijcai25}

\end{document}